\documentclass{article}

\usepackage{arxiv}

\usepackage[utf8]{inputenc} 
\usepackage[T1]{fontenc}    
\usepackage{hyperref}       
\usepackage{url}            
\usepackage{booktabs}       
\usepackage{nicefrac}       
\usepackage{microtype}      
\usepackage{amsfonts}       
\usepackage{lipsum}		
\usepackage{graphicx}
\usepackage{natbib}
\usepackage{doi}
\usepackage{verbatim}

\usepackage{amsmath}
\usepackage{amssymb}
\usepackage{mathtools}
\usepackage{amsthm}

\usepackage{booktabs}
\usepackage{multirow}
\usepackage[table]{xcolor}

\usepackage{microtype}
\usepackage{graphicx}
\usepackage{subcaption}
\usepackage{booktabs} 
\usepackage{xspace}
\usepackage{wrapfig}
\usepackage{subcaption}

\usepackage{tabularx}
\usepackage{array}
\usepackage[most]{tcolorbox}
\usepackage{fvextra}
\usepackage{xcolor}

\usepackage{tabularx}
\usepackage{array}
\usepackage[table]{xcolor}
\usepackage{booktabs}

\newcommand{\oursys}{LookAheadOPD\xspace}

\definecolor{brightmaroon}{rgb}{0.76, 0.13, 0.28}

\title{LookAheadOPD: Progress-Guided On-Policy Distillation for Agentic Tasks}

\author{{\hspace{1mm}Haojun Shi} \\
	Department of Computer Science\\
	Yale University\\
    haojun.shi@yale.edu \\
	\And{\hspace{1mm}Arman Cohan} \\
	Department of Computer Science\\
	Yale University\\
    arman.cohan@yale.edu\\
}

\date{}

\renewcommand{\shorttitle}{LookAheadOPD: Progress-Guided On-Policy Distillation for Agentic Tasks}

\begin{document}
\maketitle

\begin{abstract}
On-policy distillation (OPD) has emerged as an effective approach for transferring the capabilities of large language models to smaller students by providing dense token-level supervision on student-generated trajectories. However, in multi-turn agentic tasks, not all decisions are equally consequential, and the teacher's preferred action may not always lead to better task progress. Existing OPD methods largely rely on token-level distributional differences or teacher interventions without explicitly evaluating the downstream consequences of competing actions. We propose LookAheadOPD, a simple framework that selectively adjusts distillation strength based on short-horizon task progress. Our method first identifies decision points where the teacher and student disagree on environment-level actions, then performs short-horizon lookahead along their respective continuations to assess the utility of teacher supervision. The resulting progress estimates are used to adaptively weight the distillation objective. We evaluate LookAheadOPD on three multi-turn agentic benchmarks—ALFWorld, ScienceWorld, and WebShop—using multiple teacher–student model configurations. LookAheadOPD consistently outperforms existing OPD baselines in most experiments, with the 4B student achieving better performance than its 30B teacher. Our results highlight the value of assessing downstream task progress when allocating teacher supervision in agentic on-policy distillation.
\end{abstract}

\section{Introduction}

On-Policy Distillation (OPD) has recently emerged as a powerful post-training technique for improving LLM capabilities \citep{agarwal2024onpolicydistillationlanguagemodels}. By providing dense token-level supervision signals from an external teacher on the student model's own sampled rollouts, OPD can effectively transfer knowledge from the teacher to the student model. Following its success in domains such as mathematical reasoning and coding \citep{jin2026entropyawareonpolicydistillationlanguage, zhang2026reinforcementawareknowledgedistillationllm}, recent works have also used OPD in multi-step agentic tasks such as online shopping (WebShop, \citep{yao2023webshopscalablerealworldweb}) and household organization (ALFWorld, \citep{shridhar2021alfworldaligningtextembodied}).
Compared with RLVR methods that use binary task-outcome rewards, the dense token-level supervision of OPD provides more fine-grained gradient signals for each individual step of the student's rollout trajectories, making it an attractive training paradigm for agentic tasks \citep{zhong2026sodstepwiseonpolicydistillation, lin2026onpolicydistillationverifiablereward}.

Despite its promise, effective OPD training in multi-turn environments faces a fundamental challenge: not all decisions have equally important consequences, and the value of any individual decision is often unclear at the current step and only becomes evident through future interactions. For example, in a household task of rearranging an object whose location is unknown at the start, the steps involved in searching for the object are more crucial than those involved in grabbing the object or moving it to another location, which are just routine steps after successfully locating the object. Moreover, actions such as moving into a different room have ambiguous value at the current step; the effectiveness of such an action will only become evident after future steps involving opening containers and looking for objects within that room. This suggests that effective OPD for agentic settings should differentiate between turns: instead of imitating the teacher universally, training should identify crucial decision points where teacher guidance is valuable and assess the utility of that guidance through subsequent task progress.

Research on agentic OPD has focused on improving the reliability of teacher supervision. One line of work aims to reduce the state drift caused by weak student rollouts: the teacher model generates some turns of the rollouts used for training, preventing the accumulation of student errors that could lead to unfamiliar environment states for the teacher \citep{li2026onpolicydistillationcurriculumturnlevel, wang2026tcodexploringtemporalcurriculum, xia2026dashopddiscrepancyawareswitchinghysteresis, liu2026stochasticteacherinterventionagentic}. While teacher takeover mitigates the effects of compounding student errors, it also shifts the training trajectories away from the student's own policy, undermining the on-policy training process. Another line of work aims to apply OPD with different weights across turns based on metrics such as the teacher-student gap, teacher or student uncertainty, or the teacher's own judgment \citep{zhong2026sodstepwiseonpolicydistillation, zhou2026sageopdselectiveagentguidedintervention, tan2026atodannealedturnawareonpolicy}. However, these methods primarily make myopic decisions about the supervision strength at each turn based only on currently available information, without considering the potential impact of the action taken on future steps.

Several recent works, therefore, use future information to assess the effectiveness of teacher supervision at the current step: they create a teacher-corrected and student-continued branch and modify OPD training at the starting step based on the teacher-student gap \citep{chen2026lookaheaddistillfuture} or the final task outcome \citep{zhang2026teacherstudentgapsenoughoutcomeguided} within that branch. For the former, the teacher-student gap is just a proxy for evaluating the reliability of teacher supervision. For the latter, the effect of the original decision is confounded with subsequent errors or recoveries when continuing to the task's end. These limitations motivate evaluating teacher supervision through its potential to advance the task from the student's current state. Disagreement alone does not establish that guidance is needed, as the student may already be making progress along a valid alternative path. Likewise, student stagnation does not establish that the teacher can offer useful guidance. We therefore assess short-horizon task progress under both student and teacher continuations, assigning the strongest supervision when the student stalls but the teacher makes progress. This local assessment captures consequences that require subsequent interactions to become apparent while limiting the influence of distant errors or recoveries on the supervision signal.

Based on these considerations, we propose \oursys, a two-stage framework for selective OPD at the turn level in agentic environments. First, we introduce action-disagreement filtering to identify candidate supervision turns. We directly ask whether the teacher would choose the same environment action as the student and suppress OPD training at steps where the teacher agrees with the student. Second, at disagreement points, we perform short-horizon progress assessments over both the student and teacher continuations and combine both pieces of evidence to determine the OPD weight at the original turn. Across three agentic benchmarks, WebShop \citep{yao2023webshopscalablerealworldweb}, ALFWorld \citep{shridhar2021alfworldaligningtextembodied}, and ScienceWorld \citep{wang2022scienceworldagentsmarter5th}, \oursys consistently outperforms the agentic OPD baselines and the teacher model itself, showing the effectiveness of integrating short-horizon progress evaluation into the agentic OPD process.

To summarize, our contributions are as follows: 1. We propose \oursys, a progress-guided OPD framework that combines action-level disagreement filtering with short-horizon student and teacher lookahead to adaptively weight supervision without additional student rollouts. 2. We demonstrate consistent improvements over five agentic OPD baselines across ALFWorld, WebShop, and ScienceWorld and across two teacher-student model configurations.
Our code is available at \url{https://github.com/stonehj05/LookAheadOPD}.

\section{Related works}

\paragraph{Teacher takeover in rollout.}
Many agentic OPD methods have the teacher generate some turns in the student rollout to avoid out-of-distribution states caused by accumulated student errors, while gradually reducing the frequency of teacher takeover. Guided-OPD \citep{li2026onpolicydistillationcurriculumturnlevel} flips a weighted coin to determine whether the teacher or student generates each turn in the rollout and gradually reduces the probability of teacher generation. TCOD \citep{wang2026tcodexploringtemporalcurriculum} has the student generate the first or last $k$ turns of the rollouts and gradually increases the number of student-generated turns $k$. DASH-OPD \citep{xia2026dashopddiscrepancyawareswitchinghysteresis} switches rollout generation between the teacher and student based on the teacher-student gap. STI-OPD \citep{liu2026stochasticteacherinterventionagentic} maps the teacher-student gap to the probability of teacher intervention (a larger gap indicates a higher probability) when generating each turn of the rollouts. In all these methods, some turns in the training rollouts are generated by the teacher instead of the student, undermining the on-policy nature of the training process.

\paragraph{Myopic decisions for selective OPD.}
Other works recognize that OPD should be applied with different weights at the turn level based on the reliability of teacher supervision. SOD \citep{zhong2026sodstepwiseonpolicydistillation} compares the teacher-student gap at each turn with the gap at the initial turn and downweights turns with larger gaps than the initial turn. ATOD \citep{tan2026atodannealedturnawareonpolicy} jointly uses student uncertainty and the teacher-student gap to determine the OPD weight at each turn. SAGE-OPD \citep{zhou2026sageopdselectiveagentguidedintervention} uses teacher judgments for selective OPD, similar to \oursys. However, it asks the teacher to directly determine the intervention strength, unlike \oursys, which only asks the teacher whether it agrees with the student's proposed environment action. All these methods determine the weight of each turn based only on information available at that turn. However, in a multi-turn environment, the consequences of student actions or the utility of teacher intervention may only become clear after examining future steps.

\paragraph{Future-involved agentic OPD.}
Several recent agentic OPD methods take future information into consideration. Despite differences in implementation details, all these methods involve identifying crucial intervention points using the teacher-student gap and constructing a teacher-corrected (single-step) and student-continued branch for lookahead. FutureBridge-OPD \citep{chen2026lookaheaddistillfuture} and Frontier-OPD \citep{jiang2026frontieropddiscoveringedgeagent} utilize trends in the teacher-student gap within that branch to determine whether to include the teacher-correction step as additional training context. However, the teacher-student KL gap is only a proxy for the utility of teacher intervention: a decreasing trend in KL does not necessarily indicate that the teacher has made a successful correction. OG-OPD \citep{zhang2026teacherstudentgapsenoughoutcomeguided} executes the student-continued branch until the task ends and determines the weight for the starting turn based on the task success of the counterfactual branch. However, executing the branch to the end means that the utility of teacher supervision could be confounded by later errors or recoveries made by the student. Compared with these methods, \oursys identifies divergence points more reliably by detecting turns where the teacher actually proposes a different environment action, rather than merely having a different action distribution. It also assesses the utility of teacher supervision more robustly through local short-horizon lookahead on both the original student rollout and the counterfactual teacher continuation branch.

\section{Method}


We propose \oursys, a method that allocates teacher supervision in agentic OPD by assessing whether the student needs guidance and whether the teacher can provide useful guidance from the same state. We first collect a student-generated trajectory and ask the teacher whether it would choose the same environment action at each turn. We retain only disagreement turns as candidates for supervision, focusing on differences that could change the agent's behavior. At each candidate turn, we examine a short window of the student's subsequent interactions and generate a teacher continuation from the same environment state over the same horizon. The teacher then evaluates whether each continuation makes progress toward the task goal, with observed task outcomes used when a continuation terminates. These progress assessments determine the distillation weight at the original turn, with the strongest supervision assigned when the student stalls but the teacher makes progress. Teacher continuations are used solely to assess supervision utility and are not added to the training trajectories. Figure \ref{fig:method} provides an overview of the method.

\subsection{Preliminaries}

\paragraph{Multi-turn Interactions.}
We consider an agent interacting with an environment over a sequence of turns. At turn $t$, the agent receives an interaction history $h_t$, consisting of the task instruction and previous actions and observations. A student policy $\pi_\theta$ generates a response
$y_t \sim \pi_\theta(\cdot \mid h_t)$, from which an environment action $a_t$ is extracted. Executing $a_t$ produces a new observation $o_{t+1}$, updating the interaction history as
\begin{equation}
    h_{t+1} = (h_t, y_t, o_{t+1}).
\end{equation}
Repeating this process produces a multi-turn trajectory
$\tau = \{(h_t, y_t, a_t, o_{t+1})\}_{t=1}^{T}$,
where $T$ denotes the number of interaction turns.

\paragraph{On-Policy Distillation.}
On-policy distillation (OPD) trains a student policy $\pi_\theta$ using supervision from a fixed teacher policy $\pi_T$ on student-generated trajectories. At each turn, the teacher evaluates the student's response conditioned on the student's interaction history. The standard OPD \citep{gu2026minillmonpolicydistillationlarge} objective minimizes the reverse KL divergence between the student and teacher distributions:
\begin{equation}
    \mathcal{L}_{\mathrm{OPD}}(\theta)
    =
    \mathbb{E}_{\tau \sim \pi_\theta}
    \left[
    \frac{1}{T}\sum_{t=1}^{T}
    \mathcal{L}_{\mathrm{KL}}(h_t;\theta)
    \right],
\end{equation}
where
\begin{equation}
    \mathcal{L}_{\mathrm{KL}}(h_t;\theta)
    =
    \frac{1}{|y_t|}
    \sum_{j=1}^{|y_t|}
    D_{\mathrm{KL}}\left(
        \pi_\theta(\cdot \mid h_t,y_{t,<j})
        \,\|\, 
        \pi_T(\cdot \mid h_t,y_{t,<j})
    \right).
\end{equation}
Here, $y_{t,<j}$ denotes the response prefix preceding token $j$. Standard OPD applies teacher supervision uniformly across student-generated turns.

\subsection{\oursys}

\begin{figure}[h!]
    \centering
    \includegraphics[width=\textwidth]{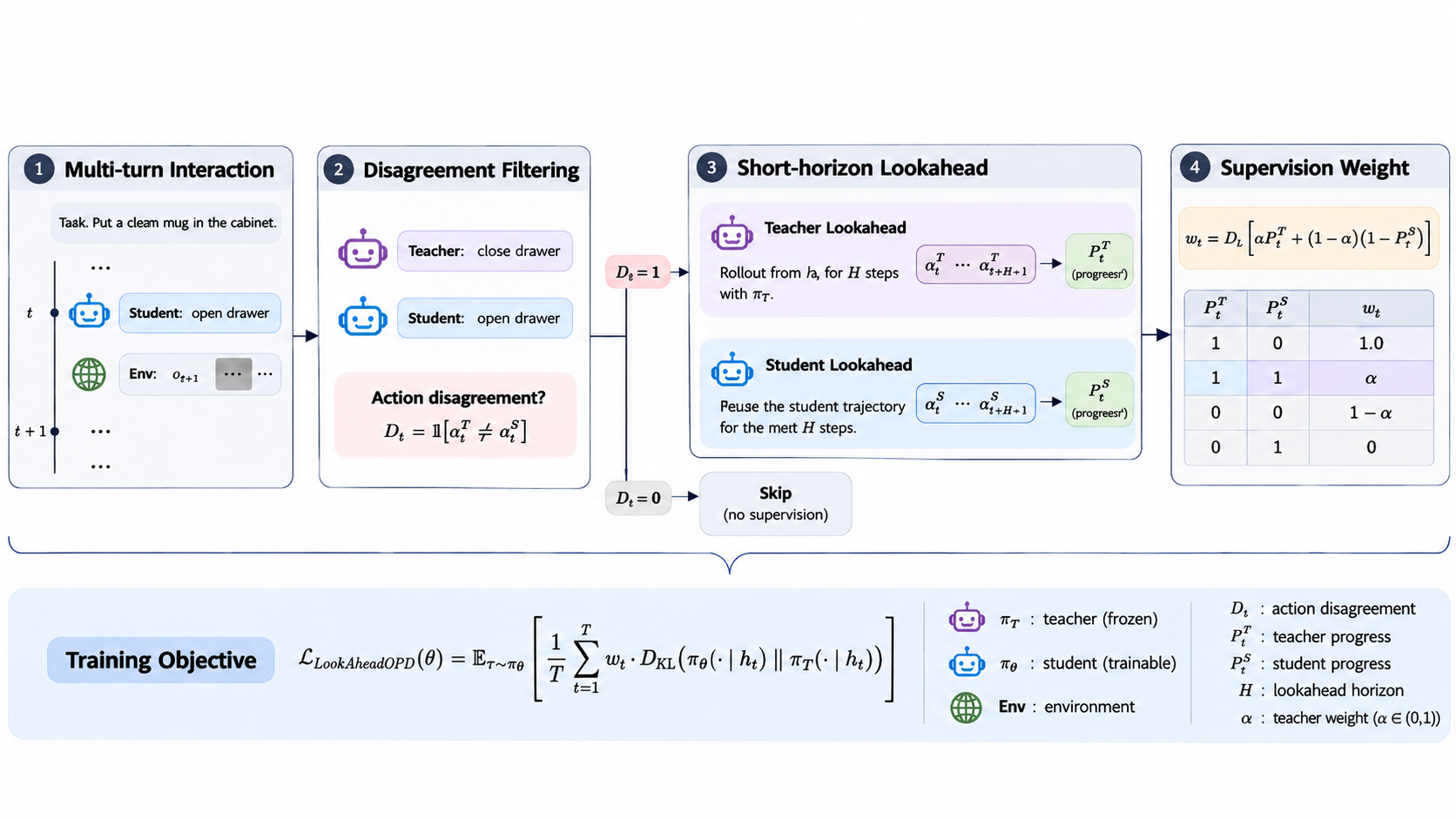}
    \caption{
        Overview of \oursys. Given a student-generated interaction trajectory, the disagreement filtering process identifies the turns where the teacher and the student disagree on the environment-level action. At these turns, we have a short-horizon task progress evaluation on both the student continuation from the existing rollout and an independently generated teacher continuation.
        The resulting progress estimates determine the
        turn-level supervision weights for OPD.
    }
    \label{fig:method}
\end{figure}

Given a student-generated trajectory $\tau$, \oursys assigns a weight $w_t$ to the turn-level distillation loss $\mathcal{L}_{\mathrm{KL}}(h_t;\theta)$. We construct this weight from an action-disagreement indicator $D_t$ and two progress indicators, $P_t^S$ and $P_t^T$, for the student and teacher continuations, respectively. We define these indicators below before presenting the resulting weighted objective.


\subsubsection{Disagreement based filtering}

We first identify candidate turns where the teacher and student disagree on the environment-level action. For each
student-generated response $y_t$, we query the teacher to
determine whether it would choose the same action $a_t$
given the same interaction history $h_t$ the student is conditioned on.

We define a binary disagreement indicator:
\begin{equation}
    D_t =
    \begin{cases}
        1, & \text{if the teacher disagrees with } a_t,\\
        0, & \text{otherwise}.
    \end{cases}
\end{equation}

When $D_t=0$, the teacher agrees with the student's action, and we exclude the turn from OPD supervision. When $D_t=1$,
we retain the turn as a candidate for further evaluation.

Unlike token-level KL for the teacher and the student's distributions, this criterion directly identifies differences in the actions the two agents would execute. In agentic environments, the actual difference in the proposed actions is much more important than the token-level KL, effectively identifying the candidate turns where the teacher supervision is potentially useful. However, 
disagreement alone does not establish the utility of the teacher supervision. We therefore assess the teacher supervision utility of both agents at the disagreement turns.

\subsubsection{Student Lookahead}
At each disagreement turn $t$, we evaluate whether the student makes meaningful progress following its proposed action. To achieve this, we reuse the subsequent $H$ turns from the already-collected on-policy trajectory to construct a short future window:
\begin{equation}
    \tau_t^S
    =
    \{(h_k,a_k,o_{k+1})\}_{k=t}^{\min(t+H-1,T)}.
\end{equation}

We provide the teacher with the task objective, the
interaction state at turn $t$, and the observed student
continuation $\tau_t^S$. The teacher assesses whether
these actions make meaningful progress toward the task goal.

This produces a binary student progress indicator:
\begin{equation}
    P_t^S =
    \operatorname{Judge}_{T}(h_t,\tau_t^S)
    \in \{0,1\},
\end{equation}
where $P_t^S=1$ indicates that the student makes progress
and $P_t^S=0$ indicates a lack of meaningful progress. When $t+H-1$ is within the total steps $T$, $P_t^S$ is obtained by the teacher assessment. The teacher judge receives the same input for the environment and the history context as well as the lookahead window. Given this information, the model directly answers the question of whether the lookahead window is making progress or not. Otherwise (lookahead reaches the task end), $P_t^S$ is just the task success. 

The student progress indicator $P_t^S$ provides information related to the student's stagnation. When the student is not stagnating and is making meaningful progress toward the task goal, the difference between the student and the teacher might be caused by the student taking a promising alternative path other than the one selected by the teacher. In such cases, the OPD supervision should be down-weighted.

\subsubsection{Teacher Lookahead}

Student stagnation alone does not establish whether the teacher supervision itself is reliable. We therefore also evaluate the teacher's ability to make progress from the same disagreement state to better establish the utility of teacher supervision.

For each candidate turn $t$, we restore the environment
to the state corresponding to $h_t$ and allow the teacher
policy $\pi_T$ to interact with the environment for up
to $H$ turns. This produces a teacher continuation:
\begin{equation}
    \tau_t^T
    \sim
    \operatorname{Rollout}(\pi_T,h_t,H).
\end{equation}

We then ask the teacher to evaluate whether its continuation
makes meaningful progress toward the task goal, obtaining
\begin{equation}
    P_t^T =
    \operatorname{Judge}_{T}(h_t,\tau_t^T)
    \in \{0,1\}.
\end{equation}

where $P_t^T = 1$ indicates that the teacher is able to make meaningful task progress in its own continuation. If the teacher continuation results in task success or failure within the $H$ steps, the task success/failure is used to determine $P_t^T$, without any teacher judgements.

Unlike previous works \citep{chen2026lookaheaddistillfuture, jiang2026frontieropddiscoveringedgeagent} involving a teacher correction branch, where they have the teacher only correct one action before the student continuation, we here use the teacher for the continuation. With student continuation, the failure within the continued branch is confounded with the inability of the student to follow through after the teacher's useful guidance.

\subsubsection{Overall training objective}

We combine action disagreement and short-horizon progress
to determine the strength of OPD supervision at each turn.

Specifically, we define the turn-level supervision weight as

\begin{equation}
    w_t =
    D_t
    \left[
        \alpha P_t^T
        +
        (1-\alpha)(1-P_t^S)
    \right],
    \label{eq:progress_weight}
\end{equation}
where $\alpha \in [0,1]$ balances evidence of teacher
progress and student stagnation. The resulting objective is
\begin{equation}
    \mathcal{L}_{\oursys}(\theta)
    =
    \mathbb{E}_{\tau \sim \pi_\theta}
    \left[
        \frac{1}{T}
        \sum_{t=1}^{T}
        w_t
        \mathcal{L}_{\mathrm{KL}}(h_t;\theta)
    \right].
    \label{eq:oursys_objective}
\end{equation}

At agreement turns, $D_t=0$ and no OPD supervision is
applied. When the teacher agrees with the actions proposed by the student, the token-level mismatch between the teacher and the student is not a useful learning signal. At disagreement turns, supervision is strongest when the teacher makes progress while the student does not. Conversely, there is no OPD supervision when the teacher is not making progress and the student is making progress, where in such cases the teacher is unable to provide useful supervision.

\section{Experiments}

\subsection{Experimental setup}

\paragraph{Benchmarks and Models.}
We evaluate \oursys on three multi-turn agentic benchmarks:
ALFWorld (household tasks), ScienceWorld (scientific experiment tasks), and WebShop (online shopping tasks). We follow the prompt templates and evaluation protocols of DASH-OPD\citep{xia2026dashopddiscrepancyawareswitchinghysteresis}, with additional details provided in the Appendix. All experiments use the Qwen3 model family \citep{yang2025qwen3technicalreport}, with two teacher--student configurations: Qwen3-30B-A3B $\rightarrow$ Qwen3-4B and Qwen3-8B $\rightarrow$ Qwen3-1.7B. We report task success rate (SR) and task score (where available), using the avg@4 evaluation protocol.

\paragraph{Training setup.} We build our implementation primarily on the Trinity-based training framework \citep{pan2025trinityrftgeneralpurposeunifiedframework}
used in the official TCOD implementation, and follow most of its training configurations. For \oursys, we set the lookahead length H = 5 from a pilot study with the Qwen3-30B-A3B teacher: on both the student and the teacher lookahead windows collected from the training set, the teacher's progress judgement with H=5 achieves 86.4\% accuracy compared to Gemini 3.8 Flash's judgements. We set $\alpha = 0.7$ to give teacher progress more weight.
Experiments with the 8B teacher / 1.7B student pair on ALFWorld and WebShop use 8 TPU v4 chips, while the remaining experiments use 16 TPU v4 chips. Further implementation details are provided in the Appendix.

\paragraph{Baselines. } We compare \oursys against vanilla OPD and four recent agentic OPD methods. \textbf{TCOD} \citep{wang2026tcodexploringtemporalcurriculum}
introduces a temporal curriculum over student-controlled rollout horizons. We adopt its backward-to-forward (B2F) variant, which progressively increases the student-controlled suffix of teacher-initialized trajectories. \textbf{UOPD} \citep{zhang2026uopduncertaintyawareinterventiononpolicy}
uses teacher uncertainty to determine whether the student
executes its proposed action or receives a teacher intervention, combining OPD with imitation-based supervision. \textbf{DASH-OPD} \citep{xia2026dashopddiscrepancyawareswitchinghysteresis}
adaptively switches between student- and teacher-controlled turns based on teacher--student policy discrepancy, applying
distillation or imitation objectives according to the
controlling policy. \textbf{FutureBridge-OPD} \citep{chen2026lookaheaddistillfuture} evaluates candidate teacher corrections using short student continuations and adds validated teacher actions as additional distillation targets. All baselines are trained under approximately matched wall-clock budgets on identical hardware within each experimental configuration. We follow the recommended hyperparameters from their official implementations.

\subsection{Main results}

\begin{table}[h!]
\centering
\footnotesize
\setlength{\tabcolsep}{3pt}
\renewcommand{\arraystretch}{1.12}

\begin{tabularx}{\textwidth}{
    @{}
    >{\raggedright\arraybackslash}p{0.25\textwidth}
    *{5}{>{\centering\arraybackslash}X}
    @{}
}
\toprule
& \textbf{ALFWorld}
& \multicolumn{2}{c}{\textbf{WebShop}}
& \multicolumn{2}{c}{\textbf{ScienceWorld}} \\
\cmidrule(lr){2-2}
\cmidrule(lr){3-4}
\cmidrule(lr){5-6}

\textbf{Method}
& SR $\uparrow$
& Score $\uparrow$
& SR $\uparrow$
& Score $\uparrow$
& SR $\uparrow$ \\
\midrule

\rowcolor{orange!12}
\multicolumn{6}{c}{
\textbf{Qwen3-30B-A3B $\rightarrow$ Qwen3-4B}
} \\

Student (zero-shot)
& 37.7 & 28.0 & 10.5 & 38.5 & 15.1 \\

Teacher (zero-shot)
& 39.7 & 40.1 & 16.0 & 52.3 & 20.9 \\

Vanilla OPD
& 38.1 & 37.7 & 15.0 & 47.2 & 18.1 \\

TCOD
& 37.7 & 35.3 & 14.8 & 46.9 & 19.8 \\

UOPD
& 33.6 & 21.4 & 9.2 & 51.0 & 17.9 \\

DASH-OPD
& 39.4 & 41.5 & 14.5 & 50.0 & 21.6 \\

FutureBridge-OPD
& 39.8 & 33.8 & 13.2 & 46.0 & 19.9 \\

\rowcolor{blue!10}
\textbf{\oursys}
& \textbf{47.4}
& \textbf{54.8}
& \textbf{22.2}
& \textbf{53.2}
& \textbf{26.2} \\

\midrule

\rowcolor{violet!10}
\multicolumn{6}{c}{
\textbf{Qwen3-8B $\rightarrow$ Qwen3-1.7B}
} \\

Student (zero-shot)
& 2.6 & 3.7 & 2.2 & 10.8 & 1.7 \\

Teacher (zero-shot)
& 42.0 & 42.1 & 17.0 & 51.0 & 25.3 \\

Vanilla OPD
& 23.9 & 39.2 & 10.5 & 10.3 & 0.0 \\

TCOD
& 27.6 & 32.7 & 13.8 & \textbf{38.8} & \textbf{11.8} \\

UOPD
& 9.7 & 53.0 & 16.0 & 35.3 & 11.7 \\

DASH-OPD
& 29.7 & 34.5 & 11.0 & 34.4 & 10.1 \\

FutureBridge-OPD
& 25.2 & 46.0 & 15.2 & 35.3 & 9.1 \\

\rowcolor{blue!10}
\textbf{\oursys}
& \textbf{31.9}
& \textbf{54.3}
& \textbf{21.0}
& 30.7
& 10.0 \\

\bottomrule
\end{tabularx}
\vspace{8pt}
\captionsetup{font=normalsize}
\caption{
Task performance of \oursys{} on ALFWorld, WebShop,
and ScienceWorld. WebShop and ScienceWorld report success
rate (SR) and task score, while ALFWorld reports SR on
its OOD test set. Higher values are better.
TCOD uses its B2F variant.
All results are evaluated using avg@4.
}
\label{tab:main_results}

\end{table}

Table \ref{tab:main_results} shows the results of the experiment. With the 4B student, \oursys achieves the best performance on all benchmarks and metrics over all the baselines. On ALFWorld, \oursys improves 7.7\% in SR over the teacher. On WebShop, \oursys improves 6.2\% in SR and 14.7 in score over the teacher. On ScienceWorld, \oursys improves 5.3\% in SR over the teacher. The consistent result that \oursys outperforms the teacher has shown that the short-horizon lookahead in \oursys has introduced more capability to the 4B student than its 30B-A3B teacher. For the 1.7B student, it still has better performance than all baselines over ALFWorld and WebShop, with improvement over the teacher model in WebShop. 

On ScienceWorld benchmark with the 1.7B student, \oursys has comparable performance to the baselines over SR while lagging behind in score. This is because of the weak 1.7B student in this benchmark: in ScienceWorld environments, the action \textit{look around} is used for examining the surroundings, while other actions like \textit{focus on} are used for actually executing actions within the environment. The 1.7B student starts in a weak state, so most of its non-\textit{look around} actions are wrong and get negative reward from the teacher supervision. However, as \textit{look around} is never an extremely wrong action, it always gets some probability in the teacher's generation, resulting in the OPD training reinforcing \textit{look around} behavior. The vanilla OPD run results in 80.8\% of actions being \textit{look around} in the end, explaining its collapse to 0 success rate. \oursys relieves this by skipping and selectively weighting the turns, where the useless repetitive \textit{look around} turns are either skipped or down-weighted, resulting in 50.5\% \textit{look around} actions (for reference, the teacher execution has 20.2\% \textit{look around}). The baselines here all have some part of teacher generation in the trained trajectory (TCOD and DASH-OPD all directly involve teacher generation, UOPD does SFT on teacher generation for uncertain turns, and FutureBridge-OPD is built on top of the TCOD-B2F variant), also relieving this \textit{look around} collapse problem.

\begin{figure}
    \centering
    \includegraphics[width=\textwidth]{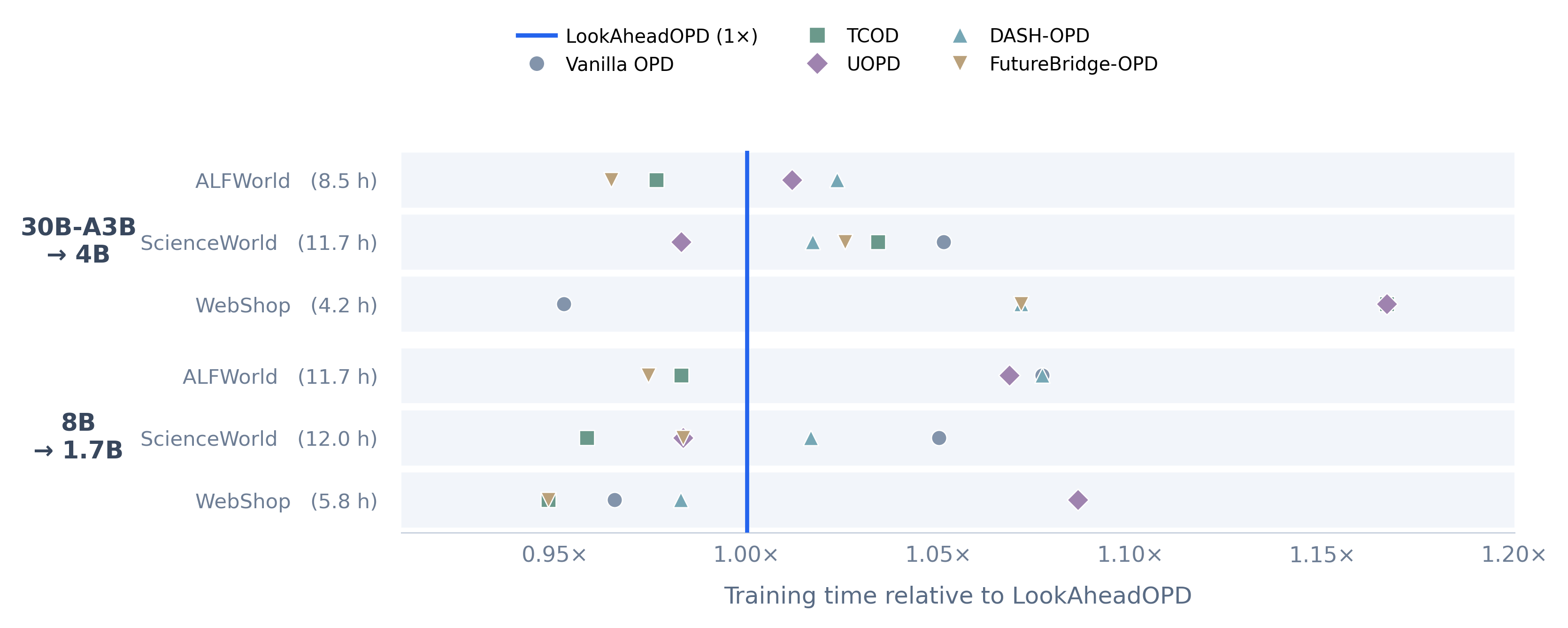}
    \caption{
        Training wall-clock time across six OPD methods
        on ALFWorld, ScienceWorld, and WebShop, using
        (a) a 30B-A3B teacher with a 4B student and
        (b) an 8B teacher with a 1.7B student.
        All methods use identical hardware within each
        benchmark and model configuration, with
        approximately matched training durations.
    }
    \label{fig:training_time}
\end{figure}

With the extra teacher call for disagreement filtering and progress evaluation, as well as constructing the teacher continuation, \oursys involves more compute per step compared with vanilla OPD. As shown in Figure \ref{fig:training_time}, all the methods have approximately matched training time on identical hardware within each model and benchmark combination to ensure fair comparison.

\clearpage

\subsection{Model Family Generalization}

\begin{wraptable}[9]{r}{0.44\textwidth}
\vspace{-12pt}
\centering
\footnotesize
\setlength{\tabcolsep}{8pt}
\renewcommand{\arraystretch}{1.1}

\begin{tabular}{lc}
\toprule
\textbf{Method} & \textbf{ALFWorld SR} \\
\midrule

\rowcolor{orange!12}
\multicolumn{2}{c}{
\textbf{Gemma4-26B $\rightarrow$ Gemma4-E2B}
} \\

Student (zero-shot) & 7.8 \\
Teacher (zero-shot) & 39.2 \\
Vanilla OPD & 26.3 \\
DASH-OPD & 25.6 \\

\rowcolor{blue!10}
\textbf{\oursys} & \textbf{34.0} \\

\bottomrule
\end{tabular}

\caption{Gemma model results.
\oursys{} outperforms Vanilla OPD and DASH-OPD.}
\label{tab:gemma_results}
\vspace{-10pt}
\end{wraptable}

To test the generalizability of \oursys{} to different model families, we evaluate it on the Gemma4 model family, using
Gemma4-26B as the teacher and Gemma4-E2B as the student.
We compare our method against Vanilla OPD and DASH-OPD on ALFWorld. As shown in Table~\ref{tab:gemma_results}, our method outperforms both baselines, suggesting that its benefits extend beyond the Qwen model family. As in Table~\ref{tab:main_results}, all methods are trained under approximately matched wall-clock budgets.
\par

\vspace{10pt}

\subsection{Ablation experiments}
To verify that the proposed short-horizon lookahead actually pays off, we conduct an ablation study with the Qwen3-30B-A3B teacher and the Qwen3-4B student. The ablation run removes both the student and teacher-side lookahead and just keeps the disagreement filtering. Table \ref{tab:component_ablation} shows the result of the ablation run. Without the lookahead, the performance across all three benchmarks has all significantly dropped, testifying that selective turn-level reweighting of OPD supervision strength from the lookahead has significant effect over the agentic OPD training.

\begin{table}[h]
\centering
\footnotesize
\setlength{\tabcolsep}{3pt}
\renewcommand{\arraystretch}{1.12}

\begin{tabularx}{\textwidth}{
    @{}
    >{\raggedright\arraybackslash}p{0.25\textwidth}
    *{5}{>{\centering\arraybackslash}X}
    @{}
}
\toprule
& \textbf{ALFWorld}
& \multicolumn{2}{c}{\textbf{WebShop}}
& \multicolumn{2}{c}{\textbf{ScienceWorld}} \\
\cmidrule(lr){2-2}
\cmidrule(lr){3-4}
\cmidrule(lr){5-6}

\textbf{Variant}
& SR $\uparrow$
& Score $\uparrow$
& SR $\uparrow$
& Score $\uparrow$
& SR $\uparrow$ \\
\midrule

Student (zero-shot)
& 37.7 & 28.0 & 10.5 & 38.5 & 15.1 \\

Teacher (zero-shot)
& 39.7 & 40.1 & 16.0 & 52.3 & 20.9 \\

w/o lookahead
& 42.0
& 45.4
& 17.8
& 44.1
& 20.3 \\




\rowcolor{blue!10}
\textbf{\oursys{} (full)}
& \textbf{47.4}
& \textbf{54.8}
& \textbf{22.2}
& \textbf{53.2}
& \textbf{26.2} \\

\bottomrule
\end{tabularx}

\vspace{8pt}
\captionsetup{font=normalsize}
\caption{
Ablation results. The student, teacher, and full
\oursys{} results are identical to those in
Table~\ref{tab:main_results}.
The ablation removes both student and teacher
lookahead, retaining only disagreement-based filtering.
Performance decreases across all three benchmarks,
demonstrating the contribution of short-horizon
lookahead to effective supervision.
}
\label{tab:component_ablation}

\end{table}

\subsection{Qualitative analysis}

We look into the evaluation records of the 4B student and compare \oursys with the baselines for the three datasets. In general, \oursys's lookahead keeps the student from inheriting the failure of the teacher model.

\paragraph{ALFWorld.} \oursys has 47.4\% success rate over FutureBridge-OPD's 39.8\%. The advantage comes mainly from several types of tasks: for example, on \textit{heat} tasks, we have a success rate of 67.4\% vs 15.2\% over the FTB baselines. The correct way to execute the task is to directly act \textit{heat the object} when the agent is holding the object and is close to the microwave. The teacher model, however, chooses to put the object in the microwave first and then heat, which would cause the failure. Most baselines inherit that behavior. However, for \oursys, as the lookahead suggests that the incorrect way to heat the object does not make meaningful progress toward the task, it downweights the teacher supervision and thus prevents the student from learning that wrong behavior, explaining the student model's success. Similar patterns appear in other types of tasks that require knowledge of the environment, where \oursys successfully makes the student avoid the errors made by the teacher. This also explains the reason why the trained student has better performance than the teacher.

\paragraph{WebShop.} \oursys has a score of 54.8 and SR of 22.2\% over DASH-OPD (the best-performing baseline)'s 41.5 and 14.5\%. The advantage mainly comes from more purchases: \oursys makes a purchase 94\% of the time, versus DASH-OPD's 69\%, the vanilla student's 47\%, and the vanilla teacher's 66\%. In WebShop, the task without a purchase receives zero score and results in loops of searching and navigating. The baseline mostly replicates this pattern from the teacher. However, as the teacher's choice of not making a purchase and making repeated navigation would not make any task progress, the lookahead procedure in \oursys would count those as not making progress and downweight the misleading teacher supervision, resulting in a higher purchase rate and score than the teacher model itself.

\paragraph{ScienceWorld.} \oursys reaches a 26.2\% success rate and a 53.2 score, against UOPD's 17.9\% and 51.0. The gap is clearest on the task type \textit{test-conductivity}, where \oursys reaches 17.9\% against UOPD's 0.0\%. The task requires building a circuit to test the unknown substance and then committing to an answer by placing it in one of two boxes. The teacher keeps connecting, disconnecting, and inspecting the circuit until running out of steps, placing the substance in only 24.8\% of episodes. UOPD amplifies this error due to its SFT on teacher outputs: it places the substance in only 2.3\% of episodes. In \oursys, the look-ahead finds that the teacher's repeated rewiring makes no progress toward the task, so that supervision is downweighted and the student is not pulled into the loop. The student places the substance in 42.1\% of episodes, which explains its higher success rate than both UOPD and the teacher itself.

\section{Conclusion}
We introduce \oursys, a progress-guided OPD framework for multi-turn agentic tasks. Our method combines teacher–student action disagreement with short-horizon lookahead to identify the turns where teacher supervision is likely to be beneficial. By evaluating the progress of teacher and student continuations, \oursys adaptively adjusts the strength of the distillation objective. Experiments on ALFWorld, ScienceWorld, and WebShop demonstrate that \oursys improves over existing OPD methods. These improvements extend across different model families, while ablation experiments show the value of short-horizon lookahead. These results suggest that the effectiveness of agentic distillation depends not only on the discrepancy between teacher and student predictions, but also on whether the teacher's preferred decisions lead to meaningful downstream progress.

\section*{Acknowledgements}

We are grateful for compute support provided by Google's TRC program.

\bibliographystyle{unsrtnat}
\bibliography{references}  






\section{Appendix}

\subsection{Training and evaluation settings}

Table \ref{tab:training_settings} shows the training and evaluation settings used for \oursys. Most settings are inherited from the TCOD\citep{wang2026tcodexploringtemporalcurriculum} paper.

\begin{table}[h!]
\centering
\small
\setlength{\tabcolsep}{4pt}
\renewcommand{\arraystretch}{1.15}

\begin{tabularx}{\textwidth}{
    >{\raggedright\arraybackslash}p{0.32\textwidth}
    *{3}{>{\centering\arraybackslash}X}
}
\toprule
\textbf{Setting}
& \textbf{WebShop}
& \textbf{ALFWorld}
& \textbf{ScienceWorld} \\
\midrule

Training task pool
& 4,096 & 3,553 & 2,294 \\

Optimizer updates
& 100 & 200 & 250 \\

Rollout batch / update batch
& \multicolumn{3}{c}{16 / 64} \\

Optimizer
& \multicolumn{3}{p{0.64\textwidth}}{
\centering
AdamW; lr $=10^{-6}$;
$(\beta_1,\beta_2)=(0.9,0.999)$;
weight decay $0.01$
} \\

Learning-rate schedule / warm-up
& \multicolumn{3}{c}{Constant / none} \\

Gradient clipping
& \multicolumn{3}{c}{Global norm $1.0$} \\

Additional RL terms
& \multicolumn{3}{p{0.64\textwidth}}{
\centering
No critic, reference-policy KL penalty,
or entropy bonus
} \\

Training decoding
& \multicolumn{3}{p{0.64\textwidth}}{
\centering
Temperature $1.0$; disabled truncation;
at most 512 new tokens per turn
} \\

Training prompt limit
& \multicolumn{3}{c}{10,240 tokens} \\

Maximum interaction turns $H_{\max}$
& 15 & 30 & 30 \\

Interaction context
& \multicolumn{3}{p{0.64\textwidth}}{
\centering
Full history; truncate early turns if
exceeding the maximum prompt length
} \\

Reasoning-mode handling
& \multicolumn{3}{c}{Qwen3 thinking mode disabled} \\

Training precision
& \multicolumn{3}{c}{bfloat16} \\

Evaluation set
& 100 tasks & 134 OOD tasks & 1,308 tasks \\

Evaluation repeats
& \multicolumn{3}{c}{Four rollouts per task} \\

Evaluation batch size
& \multicolumn{3}{c}{16 tasks} \\

Evaluation decoding
& \multicolumn{3}{p{0.64\textwidth}}{
\centering
Temperature $0.4$; disabled truncation;
max prompt length 40,960;
max 4,096 new tokens per turn
} \\

Reported task metrics
& Score; SR & SR & Score; SR \\

\bottomrule
\end{tabularx}
\vspace{10pt}
\caption{
Training and evaluation settings shared by all six methods.
Each entry applies to both student scales.
``Disabled truncation'' denotes
top-$p=1$, top-$k=-1$, and min-$p=0$.
}
\label{tab:training_settings}
\end{table}

\subsection{Environment task prompt templates}

\label{sec:env-prompt}

We present the exact templates used by \oursys for both training and evaluation. Templates are adopted from DASH-OPD \citep{xia2026dashopddiscrepancyawareswitchinghysteresis}. For all the baselines, we use their own templates for training and evaluation. For all the experiments, we present the full interaction history to the model in the memory field of model inputs and only keep the most recent ones in the explicit prompt.

\newpage
\begin{tcolorbox}[
    enhanced,
    breakable,
    colback=white,
    colframe=blue!65!black,
    colbacktitle=blue!45!black,
    coltitle=white,
    title=\textbf{ALFWorld Task Prompt Template},
    fonttitle=\large,
    arc=2mm,
    boxrule=0.7pt,
    left=8pt,
    right=8pt,
    top=8pt,
    bottom=8pt
]

\begin{Verbatim}[fontsize=\small, breaklines=true]
You are an expert agent operating in the ALFRED Embodied Environment.
Your task is to: {task_description}
Prior to this step, you have already taken {step_count} step(s). Below are the most recent {history_length} observations and the corresponding actions you took: {action_history}
You are now at step {current_step} and your current observation is: {current_observation}
Your admissible actions of the current situation are:
[{admissible_actions}].

Now it's your turn to take an action.
You should first reason about the current situation.
Once you've finished your reasoning, you should choose the best admissible action for the current step and present it within <action> </action> tags.
Do not output any other text besides your reasoning and the action.
{admissible_actions} is the game's admissible commands, each in quotes and separated by \n , with help removed.
\end{Verbatim}

\end{tcolorbox}

\begin{tcolorbox}[
    enhanced,
    breakable,
    colback=white,
    colframe=blue!65!black,
    colbacktitle=blue!45!black,
    coltitle=white,
    title=\textbf{WebShop Task Prompt Template},
    fonttitle=\large,
    arc=2mm,
    boxrule=0.7pt,
    left=8pt,
    right=8pt,
    top=8pt,
    bottom=8pt
]

\begin{Verbatim}[fontsize=\small, breaklines=true]
You are an expert autonomous agent operating in the WebShop e-commerce environment.
Your task is to: {task_description}.
Prior to this step, you have already taken {step_count} step(s). Below are the most recent {history_length} observations and the corresponding actions you took: {action_history}
You are now at step {current_step} and your current observation is: {current_observation}.
Your admissible actions of the current situation are:
[
{available_actions}
].

Now it's your turn to take one action for the current step.
You should first reason about the current situation, then think which admissible action best advances the shopping goal.
Once you've finished your reasoning, you should choose the best admissible action for the current step and present it within <action> </action> tags.
Do not output any other text besides your reasoning and the action.
{available_actions} has one action per line: search[<query>] when the page has a search bar, then click[<item>] for each clickable element.
\end{Verbatim}

\end{tcolorbox}

\begin{tcolorbox}[
    enhanced,
    breakable,
    colback=white,
    colframe=blue!65!black,
    colbacktitle=blue!45!black,
    coltitle=white,
    title=\textbf{ScienceWorld Task Prompt Template},
    fonttitle=\large,
    arc=2mm,
    boxrule=0.7pt,
    left=8pt,
    right=8pt,
    top=8pt,
    bottom=8pt
]

\begin{Verbatim}[fontsize=\small, breaklines=true]
You are an expert agent operating in the ScienceWorld text environment.
{task_description}
Prior to this step, you have already taken {step_count} step(s). Below are the most recent {history_length} observations and the corresponding actions you took: {action_history}
You are now at step {current_step} and your current observation is: {current_observation}
Available action commands: [{action_templates}]
Available objects you can interact with: [{objects}]

Now it's your turn to take an action. Combine an action command with appropriate object(s) to form a valid action.
You should first reason about the current situation.
Once you've finished your reasoning, you should choose the best valid action for the current step and present it within <action> </action> tags.
Do not output any other text besides your reasoning and the action.
{action_templates} is the game's action templates (for example 'open OBJ'), with help removed. {objects} is the objects you can interact with. Both are comma-separated and quoted.
\end{Verbatim}

\end{tcolorbox}

\subsection{\oursys's prompt templates}

We also show the exact templates we use for \oursys components. For the progress evaluation, we use the same prompt for both the student and the teacher. Both the disagreement and the progress evaluation see the same environment context part as those in Section \ref{sec:env-prompt} and append the text to the prompt end. There are no task-specific prompts tailored for each benchmark.

\begin{tcolorbox}[
    enhanced,
    breakable,
    colback=white,
    colframe=red!65!black,
    colbacktitle=red!45!black,
    coltitle=white,
    title=\textbf{Disagreement Prompt Template},
    fonttitle=\large,
    arc=2mm,
    boxrule=0.7pt,
    left=8pt,
    right=8pt,
    top=8pt,
    bottom=8pt
]

\begin{Verbatim}[fontsize=\small, breaklines=true]
Now suppose the action chosen for the current step is:
{student_action}

Would you choose this exact action for the current step? First briefly reason about whether it is the best available action, then give your final decision wrapped in <answer></answer> tags -- either <answer>Yes</answer> or <answer>No</answer>. Do not output any other text besides your reasoning and the final answer.
\end{Verbatim}

\end{tcolorbox}

\begin{tcolorbox}[
    enhanced,
    breakable,
    colback=white,
    colframe=red!65!black,
    colbacktitle=red!45!black,
    coltitle=white,
    title=\textbf{Progress Evaluation Prompt Template},
    fonttitle=\large,
    arc=2mm,
    boxrule=0.7pt,
    left=8pt,
    right=8pt,
    top=8pt,
    bottom=8pt
]

\begin{Verbatim}[fontsize=\small, breaklines=true]
Based on the {window_len} step(s) shown above (steps {start_step}-{end_step}), do you think the agent is making progress toward completing the task? First briefly reason about whether those particular steps help advance the goal, then give your final decision wrapped in <answer></answer> tags -- either <answer>Yes</answer> or <answer>No</answer>. Do not output any other text besides your reasoning and the final answer.
\end{Verbatim}

\end{tcolorbox}

\subsection{\oursys metrics}

Table \ref{tab:gate} shows the metrics within \oursys's training runs. Table \ref{tab:gate-stats} shows the rate at which each individual component of \oursys is applied during the training. The disagreement rate indicates the portion of student turns whose actions are rejected by the teacher. The student and the teacher progress rates indicate the portion of the student or the teacher making progress in the lookahead window from the teacher's judgments. Table \ref{tab:gate-weights} shows the portion of different OPD supervision rates applied for the turns after the disagreement filtering (Figure \ref{fig:method} shows the meaning of each supervision weight). Across different benchmarks and teacher/student combinations, all of these metrics vary a lot, suggesting that \oursys is more than simply filtering out/downweighting a certain portion of turns or downweighting OPD supervision to certain weights.

\begin{table}[h!]
\centering
\small

\begin{subtable}{\linewidth}
\centering
\begin{tabular}{llccc}
\toprule
Teacher $\rightarrow$ Student & Environment & Disagreement (\%) & Student progress (\%) & Teacher progress (\%) \\
\midrule
\multirow{3}{*}{30B-A3B $\rightarrow$ 4B}
 & ALFWorld     & 24.6 & 53.6 & 79.8 \\
 & ScienceWorld & 46.9 & 29.1 & 45.8 \\
 & WebShop      & 39.6 & 35.2 & 58.0 \\
\midrule
\multirow{3}{*}{8B $\rightarrow$ 1.7B}
 & ALFWorld     & 32.3 & 22.7 & 46.9 \\
 & ScienceWorld & 52.4 &  6.0 & 14.4 \\
 & WebShop      & 66.3 & 32.6 & 68.0 \\
\bottomrule
\end{tabular}
\caption{The relevant rate for each of \oursys's components. Disagreement rate: the rate of the teacher disagreeing with the student's proposed action; student progress rate: the rate that the student lookahead over the collected rollout makes progress; teacher progress rate: the rate that the teacher lookahead over the teacher continuation makes progress.}
\label{tab:gate-stats}
\end{subtable}

\vspace{0.8em}

\begin{subtable}{\linewidth}
\centering
\begin{tabular}{llcccc}
\toprule
& & \multicolumn{4}{c}{Share of disagreement turns (\%)} \\
\cmidrule(lr){3-6}
Teacher $\rightarrow$ Student & Environment & $w=1.0$ & $w=0.7$ & $w=0.3$ & $w=0.0$ \\
\midrule
\multirow{3}{*}{30B-A3B $\rightarrow$ 4B}
 & ALFWorld     & 33.4 & 46.4 & 13.0 &  7.2 \\
 & ScienceWorld & 28.1 & 17.7 & 42.8 & 11.4 \\
 & WebShop      & 30.8 & 27.2 & 34.1 &  8.0 \\
\midrule
\multirow{3}{*}{8B $\rightarrow$ 1.7B}
 & ALFWorld     & 29.0 & 18.0 & 48.3 &  4.7 \\
 & ScienceWorld & 11.2 &  3.1 & 82.8 &  2.8 \\
 & WebShop      & 40.2 & 27.8 & 27.1 &  4.8 \\
\bottomrule
\end{tabular}
\caption{OPD supervision strength across the disagreed turns. Weight 1.0: the teacher makes progress while the student does not; weight 0.7: both the teacher and the student make progress; weight 0.3: both the student and the teacher do not make progress; weight 0.0: the student makes progress while the teacher does not.}
\label{tab:gate-weights}
\end{subtable}

\caption{Metrics of \oursys's run across the entire training process. For all the metrics, including disagreement rates, student/teacher progress rates, and the OPD supervision weights, they vary a lot across different dataset/model combinations, suggesting that \oursys is beyond simply filtering out a certain portion of turns or reweighting the OPD by a fixed weight. }
\label{tab:gate}
\end{table}

\end{document}